# An Improved Approach of Intention Discovery with Machine Learning for POMDP-based Dialogue Management

*RUTURAJ Rajendrakumar RAVAL*
raval115@uwindsor.ca
University of Windsor

## Abstract

An Embodied Conversational Agent (ECA) is an intelligent agent that works as the front end of software applications to interact with users through verbal/nonverbal expressions and to provide online assistance without the limits of time, location, and language. To help to improve the experience of human-computer interaction, there is an increasing need to empower ECA with not only the realistic look of its human counterparts but also a higher level of intelligence. This thesis first highlights the main topics related to the construction of ECA, including different approaches of dialogue management, and then discusses existing techniques of trend analysis for its application in user classification. As a further refinement and enhancement to prior work on ECA, this thesis research proposes a cohesive framework to integrate emotion-based facial animation with improved intention discovery. In addition, a machine learning technique is introduced to support sentiment analysis for the adjustment of policy design in POMDP-based dialogue management. The proposed research work is going to improve the accuracy of intention discovery while reducing the length of dialogues.

## 1. Introduction

In the world of *Natural Stupidity*, we need an *Artificial Intelligence*, therefore we require a system to incorporate these advanced technologies in our day-to-day life, that means, people's life should be less stressful to get things done with the help of available technological resources like HCI (Human-Computer Interaction) and an interaction with an ECA (Embodied Conversational Agent) and this thesis directs the work in the direction of more advanced work done using HCI and ECA as a tool to involve the POMDP and deep reinforcement learning to discover user's intention and to improve the accuracy from the user's input.

### 1.1 Introduction to Human-Computer Interaction (HCI)

Human-Computer Interaction (HCI) is a field of research within the design and use of computer technology, focused on the interfaces between people (users) and computers. Researchers in the field of HCI both observe the ways in which humans interact with computers and design technologies that let humans interact with computers in novel ways. As a field of research, human-computer interaction is situated at the intersection of [1],

- Computer Science
- Behavioural Sciences
- Design

The recent advances in cognitive psychology and related sciences lead us to the conclusion that knowledge of human cognitive behaviour is sufficiently advanced to enable its applications in computer science and other practical domains. The goal is to help to create an applied information-processing psychology, as with all applied science, this can be only done by working within some specific domain of application. This domain is called Human-Computer Interface for the Interaction. The entire world is amid transforming itself to use the power of computers throughout its entire fabric-wherever information is used and that transformation depends critically on the quality of human-computer interaction [2].

### 1.2 Introduction to Embodied Conversational Agent (ECA)

The ECAs in artificial intelligence, also sometimes referred to as an interface agent [3], is an intelligent agent that interacts with the environment through a physical body within that environment. Agents that are represented graphically with a body, for example, a human or a cartoon animal, are also called embodied agents, although they have only virtual, not physical, embodiment. A branch of AI focuses on empowering such agents to interact autonomously with human beings and the environment.

## 2. Literature review

For HCI, making Artificial Intelligence (AI) more human [4] is necessary, that means to conduct HCI between ECA and a user to make the conversation real enough, AI is required to shape the intelligent ECA. The parameters to include for this progress, involves environmental influence, agent parameters, generality, flexibility, learning speed, effect of cheap talk, loyalty, honesty, speech profile, etc. and then turing can be conducted over the agents to test the intelligence of the ECA against user-to-user for more advanced ECA-to-user conversation.

In the process of automation, two aspects are important to be considered and those are, *appearance* and *intelligence*, that how effective the ECA can be developed for the ease of access for the replica of human face-to-face interaction in terms of HCI to make it more reliable and more real. That can be decided by the animated speech, lip movement and facial expressions, eye, head and body movements to the gestures, emotion expression, etc. and last but not the least the automation process involves NLU and NLP to take an optimized decision from the ECA's end for the user that can be useful in detecting the intention of the user which analyze the understanding of what users aim or plan or actually intended about the requirement and to find out, agent needs to be intelligent enough and that is called as an *intention discovery*. With the help of intention which is discovered by the agent, what user meant, then interaction begins in a true sense as agent will be exchanging the words based on the discovery of user's aim and that focus will be carried forward when a user and an agent interacts and the number of conversations it takes for user and agent to reach the desired result of a user, is known as a *dialogue length* and to improve the policy, there is an importance of improving the optimization level of dialogue that how minimal or the shortest length it took for an agent to reach the goal and that dialogue length can be optimized by anticipating the user knowledge level. The user knowledge level can be categorized in different categories (Different knowledge level selectors to decide the policy based on the understanding of user), such as expert, professional, amateur, and novice, that how advanced the user is or how basic knowledge a user must carry forward the interaction and that will decide the policy structure of interaction with the help of Dialogue Management (DM).

### 2.1 Embodied Conversational Agent with Dialogue Management (DM)

ECAs have the same ability as per the human in terms of the representation properties as face-to-face conversation which includes, an ability to recognize and respond to the input, ability to generate an output, use of conversational functions are utilized under HCI branch. So the motivation works in terms of the interaction, the included attributes are, intuitiveness, redundancy and modality switching, the social nature of the interaction, etc. [6] The input to the DM is the human utterance, usually converted to some system-specific semantic representation by the Natural language understanding (NLU) component. The DM usually maintains some state variables, such as the dialogue history, the latest unanswered question, etc., depending on the system [7]. That is how we can combine the great features of ECA with Dialogue Manager to iterate the functionalities of a communicative artificial intelligent agent to satisfy the aim of this system.

The decision-making approach in DM world, how to reach a goal and how the decisions are made based on the data we have, there have been plenty of research work has been carried out in terms of Breadth First Search (BFS), Depth First Search (DFS) and A* algorithms in the deterministic environment where actions are predictable. These approaches work when we have a finite set of data, but it can't work under uncertainty to-predict-the-future-goals environment. Various approaches are proposed since the year 1960 to understand and model the decision-making task under uncertainty like, Markov Decision Process (MDP), etc. and ECA cannot help in discovering the intention we have taken Partially Observable Markov Decision Process (POMDP) into an action which is an extension of an MDP.

### 2.2 Dialogue management

Dialogue management is a system consisting of a dialogue manager, which is a core of a Spoken Dialogue System (SDS) with its main features like, tracking dialogue states and maintaining a dialogue policy which decides how the system reacts on given dialogue state. Recently, different approaches in automatic dialogue management policy optimization are there among the Reinforcement Learning (RL) and the POMDP has been the most famous one [8]. Statistical approaches to dialogue modelling allow automatic optimization of the SDS. An SDS is typically designed according to the structured-ontology [9]. Instead of letting a human expert write a complex set of decision rules, it is more common to use *reinforcement learning*. The dialogue is represented as a *Markov Decision Process (MDP)* - a process where, in each state, the *DM* has to select an action, based on the state and the possible rewards from each action. In this setting, the dialogue author should only define the reward function, for example: in tutorial dialogues, the reward is the increase in the student grade; in information seeking dialogues, the reward is positive if the human receives the information, but there is also a negative reward for each dialogue step. RL techniques are then used to learn policy, for an example, what kind of observation and feedback should we use in each state? etc. This policy is later used by the DM in real dialogues.

To learn an optimum policy conducted by an agent by maximizing its cumulative reward. One of the advantages of RL-based dialogue management (DM) is the robustness. RL algorithms are mostly-data demanding, which leaves dialogue system developers in worry as there are usually few or even no data available at the early stage of development. Several methods have been proposed to mitigate this problem. A user simulator is often built using wizard-of-oz dialogue data, and then the simulator is used to train an RL-based DM. In recent studies, it has been shown that by incorporating domain knowledge into the design of kernel functions [8].

The SDS enables HCI via speech so that the DM has two aims to maintain the dialogue state based on the current spoken language understanding input and the conversation history, choose a response according to its dialogue policy. To offer the robustness and to track the distribution of all the dialogue states at every dialogue turn, called as a belief state. Then the system response is based on the belief state rather than an inaccurate estimate of most likely dialogue state [10].

## 2.3 Markov Decision Process (MDP)

Markov Decision Process (MDP) is an output of the continuous cast of Dialogue management (DM) which is composed of a finite set of actions, a continuous multivariate belief state space and a reward function [11]. A Markov decision processes (MDPs) is a discrete time stochastic control process. It provides a mathematical framework for modelling decision making in situations where outcomes are partly random and partly under the control of a decision maker. MDPs are useful for studying optimization problems solved via dynamic programming and reinforcement learning [12].

At each time step, the process is in some state *s*, and the decision maker may choose any action *a* that is available in state *s*. The process responds at the next time step by randomly moving into a new state *s'*, and giving the decision maker a corresponding reward $R_a(s, s')$. The probability that the process moves into its new state *s'* is influenced by the chosen action. Specifically, it is given by the state transition function $P_a(s, s')$. Thus, the next state *s'* depends on the current state *s* and the decision maker's action *a*. But given *s* and *a*, it is conditionally independent of all previous states and actions; in other words, the state transitions of an MDP satisfies the Markov property. Markov decision processes are an extension of Markov chains; the difference is the addition of actions (allowing choice) and rewards (giving motivation). Conversely, if only one action exists for each state (e.g. "wait") and all rewards are the same (e.g. "zero"), a Markov decision process reduces to a Markov chain. A Markov decision process is a 5-tuple *(S, A, $P_a$, $R_a$, γ)*, where

- *S* is a finite set of states,
- *A* is a finite set of actions ($A_s$ is the finite set of actions available from state *s*),
- $P_a(s, a, s') = P_r(s_{t+1} = s' \mid s_t = s, a_t = a)$ is the probability that action *a* in state *s* at time *t* will lead to state *s'* at time *t+1*,
- $R_a(s, a, s')$ is the immediate reward (or expected immediate reward) received after transitioning from state *s* to state *s'*, due to action *a*,
- *γ ∈ [0, 1]* is the discount factor, which represents the difference in importance of future rewards and present rewards.

### 2.3.1 Shortcomings of MDPs

The core problem of MDPs is to find a "policy" for the decision maker: a function $\pi$ that specifies the action $\pi(s)$ that the decision maker will choose when in state *s*. Once a Markov decision process is combined with a policy in this way, this fixes the action for each state and the resulting combination behaves like a Markov chain. The goal is to choose a policy $\pi$ that will maximize some cumulative function of the random rewards, typically the expected discounted sum over a potentially infinite horizon:

$$\sum_{t=0}^{\infty} \gamma^t \, R_{a_t}(s_t, \, s_{t+1}), \text{ where at} = \pi(s_t)$$

*Equation 1 Decision maker in an MDP to find Policy*

where γ is the discount factor and satisfies $0 \leq \gamma < 1$. (For example, $\gamma = 1/(1 + r)$ when the discount rate is *r*.) γ is typically close to 1. Because of the Markov property, the optimal policy for this problem can indeed be written as a function of *s* only, as assumed above. MDP is fully observable in stochastic environment actions are random.

Therefore, the solution could be termed as a Partially Observable Markov Decision Process (POMDP) that would solve the optimization problem of MDP in the stochastic environment. Suppose we *know* the state transition function *P* and the reward function *R*, and we wish to calculate the policy that maximizes the expected discounted reward. The standard family of algorithms to calculate this optimal policy requires storage for two arrays indexed by state: *value V*, which contains real values, and *policy $\pi$* which contains actions. At the end of the algorithm, $\pi$ will contain the solution and *V(s)* will contain the discounted sum of the rewards to be earned (on average) by following that solution from state *s*.

### 2.4 Partially Observable Markov Decision Process (POMDP)

The above-given equation assumes that the state *s* is known when action is to be taken; otherwise $\pi(s)$ cannot be calculated. When this assumption is not true, the problem is called a partially observable Markov decision process or POMDP. A partially observable Markov decision process (POMDP) is a generalization of a Markov decision process (MDP). A POMDP models an agent decision process in which it is assumed that the system dynamics are determined by an MDP, but the agent cannot directly observe the underlying state. Instead, it must maintain a probability distribution over the set of possible states, based on a set of observations and observation probabilities, and the underlying MDP. The POMDP helps to build the discrete-time relationship between an agent and its environment. Formally, a POMDP is a 7-tuple *(S, A, T, R, Ω, O, γ)*, where

- *S* is a set of states: The input which is divided into a finite set of possible states
- *A* is a set of actions: A finite set of possible ACTIONS available and actions are information-driven or goal-driven
- *T* is a set of conditional transition probabilities between states: It captures the probabilistic relationship between the states and the actions executed to change the state of the world
- *R* is the reward function: It gives the relative measure of desirability to be in a state
- Ω is a set of observations: It captures the probabilistic relationship between the state and observations
- *O*, is a set of conditional observation probabilities: a finite set of observations of the state
- *γ*, is the discount factor: The discount factor decides how much immediate rewards are favoured over future rewards.

In a POMDP we add a set of observations to the model. So instead of directly observing the current state, the state gives us an observation which provides a hint about what state it is in. The observations can be probabilistic; so we need to also specify an observation function. This observation function simply tells us the probability of each observation for each state in the model. We can also have the observation likelihood depend on the action if we like [13].

## 3. Detailed Partially Observable Markov Decision Process model

### 3.1 Belief State (BS)

Belief-State is a probability distribution over all possible states which gives as much information as the entire action-observation history [14].

BS along with transition and observation probabilities helps to transform the problem from partially-observable to completely observable. An agent needs to update its belief upon taking the action *a* and observing *o*. Since the state is Markovian, maintaining a belief over the states solely requires knowledge of the previous belief state, the action taken, and the current observation.

The operation is denoted *b' = τ(b, a, o)*. Below we describe how this belief update is computed. After reaching *s'*, the agent observes *o* ∈ Ω with probability *O(o | s', a)*. Let *b* be a probability distribution over the state space *S*. *b(s)* denotes the probability that the environment is in state *s*. Given *b(s)*, then after taking action *a* and observing *o*,

$$b'(s') = \eta O(o \mid s', a) \sum_{s \in S} T(s' \mid s, a) b(s)$$

*Equation 2 Belief update*

Where **η = 1/ Pr(o | b, a)** is a normalizing constant with

$$\Pr(o \mid b, a) = \sum_{s' \in S} O(o \mid s', a) \sum_{s \in S} T(s' \mid s, a) b(s)$$

*Equation 3 Normalizing constant evaluation*

If we are given a belief state for time t and we perform an action *a* and get observation *o* we can compute a new belief state for time *t+1* by simple applying Bayes' rule and using the model parameters [13].

Bayes' theorem (conditional probability):

*P(A | B) = P(B | A) P(A) / P(B)*

*Equation 4 Bayes' theorem*

P(A) and P(B): The probabilities of observing A and B exclusively

P(A | B): The probabilities of observing event A given that B is true

P(B | A): The probabilities of observing event B given that A is true

## 3.2 Policy

In the planning of optimizing the dialogue length, the interaction system constructs a tree with the possible states and actions. By traversing the tree from the root node to leaves, the optimal plan is computed. In POMDP, as the outcomes of acts taken are stochastic, in other words as the branching factor is high, the tree constructed using conventional planning is very deep.

Policy: Belief-State → Action

**π: b(s') → a**

*Equation 5 Policy function*

In general, while an MDP policy mapped states to actions, an optimal POMDP policy maps belief states to actions. The focus is that the space of all belief states is continuous. This is a big part of why these problems are hard to solve. Nevertheless, there are algorithms that can work in the space and yield optimal solutions: though they are somewhat complex and computationally inefficient. Once the policy has been computed (optimal or otherwise), using the solution is relatively simple and computationally easy. The way in which one would use a computed policy is to start with belief about where you are in the world. Then continually [15]:

1. Use the policy to select an action for current belief state;
2. Execute the action;
3. Receive an observation;
4. Update the belief state using current belief, action and observation; then
5. Repeat.

Again, holding tight with the flow of the paper (Toward effective dialogue management using partially observable Markov decision processes) [16] [5] to take a thorough look at the policy in the DM context can be shown here. A policy is a function:

$$\pi(b) \longrightarrow a$$

*Equation 6 Policy function*

where *b* is a belief state and *a* is the action chosen by the policy *π*. An optimal policy *π\** is a policy that maximizes the expected cumulative reward:

$$\pi^* = argmax_\pi E\left[\sum_{t=0}^{\infty} \gamma^t R_t\right]$$

*Equation 7 Optimal policy*

where $R_t$ is the reward when the agent follows policy $\pi$. We define value functions $V_i: B \rightarrow R$. $V_n(b)$ is the maximum expected a cumulative reward when the agent has n remaining steps to go. Its associated policy is denoted by $\pi_n$. When the agent has only one remaining step to go (i.e. $n = 1$), all it can do is to select an action and send it to the environment.

### 3.3 Shortcomings of POMDP

The first shortcoming of POMDP is that it contains an approximation: we are trying to find representations of the belief which are rich enough to allow good control, but which are also sufficiently parsimonious to make the planning problem tractable.

The second disadvantage is a technical one: while making a nonlinear transformation of the belief space, POMDP planning algorithms which assume a convex value function will no longer work.

For non-trivial modelling, the state/action/observation spaces quickly grow, this is a result of requiring a complete enumeration of all possible states. The discrete, uniform time elapsing between each action-observation pair is not always a good assumption. The assumption that the model will not change over time (non-stationary process) does not always apply.

### 3.4 The Contextual Control Model (COCOM)

The COCOM model [17] is based upon cognitive modes. This model suggests that the system needs to decide what actions to take according to the context of the situation. There are four modes of operations strategic, tactical, opportunistic, scrambled. Each control mode has its own characteristics and type performance and the mode of team behaviour varies in terms of the degree of planning.

- Strategic mode - this mode concentrates on long-term planning using a global view and has a higher level of control. The amount of information sought, and user-system coordination required is expected to be extensive.
- Tactical mode - The system and the amount of available information is beyond what is immediately observable but may be limited to what the routine procedure requires.
- Opportunistic mode - The next action is predictable depending upon a current situation where available information is inadequate due to the less effective planning, limited time and incomplete understanding of context.
- Scrambled mode - The choice of next action is basically irrational or completely unpredictable. The type of performance is thus paradoxically characterized by the lack or absence of any control.

### 3.5 Belief state History and Trend Analysis

#### 3.5.1 POMDP-based Dialogue management

Based on the shortcomings of the POMDP approach, the approach of belief state history and trend analysis approach helps in working with Belief State by using on the HISTORY of belief states and the dynamics of Belief State (BS) [18]. The consequence is inflexibility for human-robot interaction as in the FSM-based approach, incapable of handling any ambiguity as in the frame/Bayes/MDP-based approaches, and insufficiency in dealing with uncertainties as in the POMDP-based approach. To overcome the shortcomings of handling the uncertainty while retaining the advantages of the current POMDP-based approach, this paper proposes a modified planning strategy as illustrated below.

$$\pi_{new}: \mathcal{I}'_{k-1} \cup \mathcal{I}_k \rightarrow U$$

*Equation 8 Modified planning strategy*

Both $I_k$ and $I'_{k-1}$ are still in the form of a belief state, and updating still uses the existing POMDP model. However, the addition of $I'_{k-1}$ in the modified approach, this state, introduces an important element to dialogue management, i.e., the history of belief state or the dynamics of belief state since the inception of the interaction for current the conversation. Although the history information of observations and actions is not maintained explicitly in $I'_{k-1}$, the union $I_k$ and $I'_{k-1}$ diminish the negative effect of the Markov Assumption and allows POMDP-based dialogue management to plan for actions with not only the current belief state but also the updated history before reaching the current state. At each stage of dialogue, the new approach uses the domain knowledge and constraint database to help to validate the change of belief state. A failed validation results in a roll-back of belief state to the previous state.

### 3.6 Trend analysis using Belief state

Trend analysis is the widespread practice of collecting information and attempting to spot a pattern. Although trend analysis is often used to predict future events, it could be used to estimate uncertain events in the past using the different states which we have stored inside the POMDP model. Trend analysis often refers to techniques for extracting an underlying pattern of behaviour in a time series which would otherwise be partly or nearly completely hidden by noise. If the trend can be assumed to be linear, trend analysis can be undertaken within a formal regression analysis, as described in trend estimation [20].

In general trend analysis is performed on historical data and time series data to predict the subject of interests for the future. The different approaches to trend analysis are as follows [21],

- Sampling- the historical data is split into training and testing datasets. The training dataset is used to develop a predictor model and its accuracy is determined using the testing dataset. Random sampling and reservoir-based sampling are example sampling methods
- Histogram- the trend is analyzed by constructing a histogram from the historical data by dividing the entire range of values into a series of intervals and then count how many values fall in each interval. Equi-depth and V-optimal are example histogram approaches
- Sketches- The frequency distribution of historical data is summarized by using hash functions. Count sketches and Count-Min sketches are example techniques
- Wavelets- Mathematical transformations are applied to transform the data into a set of wavelet coefficients representing a different level of granularity to analyze the trend

Example- A *trend analysis* is a method of analysis that allows traders to predict what will happen with a stock in the future. Trend analysis is based on historical data about the stock's performance given the overall trends of the market and indicators within the market. Trend analysis takes into account historical data points for a stock and, controlling for other factors like the general changes in the sector, market conditions, competition for similar stocks, it allows traders to forecast short, intermediate, and long-term possibilities for the stock [22]. When wavelets are used for trend analysis they are used to enhance the trend by removing the possibilities of hidden noise, by approaching the exact and accurate trend over time-series data.

### 3.7 POMDP using Belief state History information

The user interacts with the agent by providing the observation and the agent responds to the user by performing an action.

- State Estimator (SE)- It receives the observation as input from the user. SE computes the observation probabilities through NLP and updates the BS value *B(s')*
- Belief History Information Storage- The BS *B(s')* value computed in the SE is stored in the belief history information $b_{hist}$ storage module
- Trend Analysis- It receives the history of belief information $b_{hist}$ as input. Number of sharp variation points $N_{cp}$ is obtained by performing DWT on $b_{hist}$
- Knowledge level selector- It receives knowledge level *k* of the user and BS *B(s')* value as input. In the policy selector, a different set of policies are defined for users at the different knowledge level. Policy $\pi k$ is selected based on the value of *B(s')*
- Make Action- It is within the Policy selector module which receives the policy $\pi_k$ to execute. Each policy in POMDP is mapped from *B(s')* to actions. In Make an Action module, action *a* is chosen through COCOM modes

### 3.8 Wavelet Theory

In mathematics, a wavelet series is a representation of a square-integrable (real-valued or complex-valued) function by certain orthonormal series generated by a wavelet [23].

In mathematics, the continuous wavelet transform (CWT) is a formal (i.e., non-numerical)

tool that provides an overcomplete representation of a signal by letting the translation and scale parameter of the wavelets vary continuously [24].

In numerical analysis and functional analysis, a discrete wavelet transform (DWT) is any wavelet transform for which the wavelets are discretely sampled. As with other wavelet transforms, a key advantage it has over Fourier transforms is a temporal resolution: it captures both frequency and location information (location in time) [25].

Why DWT? - The DWT decomposes the signal to discrete time and provides enough information for both analysis and synthesis of the original signal. It helps in removing some of the samples of the signal by reducing the sampling rate to reduce the noise and to gain better accuracy. The time complexity is O(n). The DWT from the perspective of 1-D and 2-D can be implemented using Haar wavelets available in the JWave open source project. The DWT considers the input from the BSH database that the function assumes that input is of length $2^n$ where $n>1$. Then the next current length becomes the working area of the output array. The length starts at the half of the array size and every iteration is halved until the length becomes 1. Then we swap the arrays to perform the next iteration.

### 3.8.1 Sharp Variation points

The Belief State History (BSH) shows different fluctuation characteristics at different interactions for different users. The scales are distinguished in the wavelet transform, are based on zero-crossing points from wavelet transformation which is known as Sharp variation points [21]. We can identify the maximum sharp variation points and the minimum variation points based on the signal.

## 3.9 Fuzzy Logic

Fuzzy logic is a form of many-valued logic in which the truth values of variables may be any real number between 0 and 1. It is employed to handle the concept of partial truth, where the truth value may range between completely true and completely false. By contrast, in Boolean logic, the truth values of variables may only be the integer values 0 or 1 [26].

Why Fuzzy logic?

- A fuzzy logic system can be defined as the non-linear mapping of an input dataset to a scalar output data. It consists of 4 components such as,
    - Fuzzifier
    - Rules
    - Inference engine
    - Defuzzifier

The rules for the fuzzy system for this thesis can be organized as written below.

| *FAMM (Fuzzy Associative Memory Matrix)* | *Fuzzy RULE base selection based on Trend Analysis performed on Belief-state history using COCOM* | | | |
|---|---|---|---|---|
| *Sentiment Analysis Reward* | Strategic | Tactical | Opportunistic | Scrambled |
| Negative | Disgust | Anger | | Fear |
| Neutral | Fear | Sad | Surprise | Sad |
| Positive | Happy | | Surprise | |

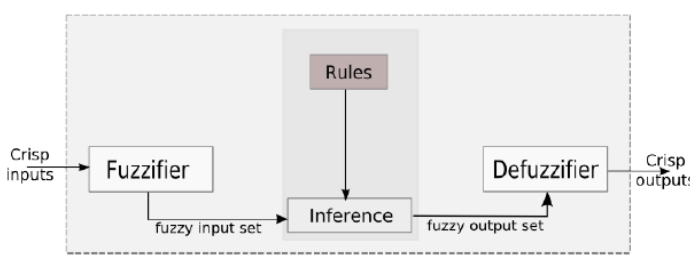

*Figure 1 Fuzzy Logic System*

## 3.10 Shortcomings of previous approaches (Significant Problems)

The general scenario of shortcomings in recently existing work can be stated as,

- POMDP refrains to capture the history of actions taken and observations made
- Dialogue length can be optimized
- POMDP hasn't been tested on a huge scale

- The policy is not accurate enough which means belief state tracing is not performed well
- The connection between different belief states and its history is not capable enough to be merged as an agent will count the different approach of same communication as two different DM context and then dialogue length will not be optimized
- Most of the models do not offer the NLP in the dialogue exchange, so that means deep learning or machine learning or artificial intelligence should be able to incorporate POMDPs
- If NLP is taken into consideration, yet patterns are found after data mining that can optimize the dialogue length and can help in improving the accuracy with better training corpus.

## 4. Proposed method

### 4.1 Sentiment Analysis

The process of computationally identifying and categorizing emotion(s) (sentiments) expressed from a piece of text. It aids in the detection of emotion.

Sentiment analysis helps in achieving end-to-end task completion and it has wide appeal as providing information about the subjective dimension of texts which can be regarded as a classification technique, either binary (polarity classification into positive/ negative) or a multi-class categorization (negative/ neutral/ positive). Most approaches use sentiment lexicons as a component (sometimes only the component). Lexicons can either be general purpose or extracted from a suitable corpus, such as movie reviews with explicit ranking information [28] [29].

### 4.2 Reinforcement Learning

The goal of Artificial Intelligence is to produce fully autonomous agents that interact with their users (environments) to learn optimal behaviours, improving over time through trial-and-error [30]. A principle mathematical framework for experience-driven autonomous learning is known as Reinforcement Learning [31]. There have been several limitations of several traditional RL approach like memory complexity, computational complexity, scalability [32]. The optimization of statistical dialogue managers using RL methods is an active and promising area of research. In contrast with the traditional discrete action domains like the Atari game (to predict the next move, from self-learning of previous states and moves, part of reinforcement learning), playing has much focus on deep RL research.

The perceptron-action-learning loop, at time $t$, the agent receives state $s_t$ from the environment, the agent uses its policy to choose an action $a_t$. After action execution, environment transitions a step, providing the next step $s_{t+1}$ as well as the feedback in the form of a reward $r_{t+1}$. The agent uses knowledge of state transitions, of the form $(s_t, a_t, s_{t+1}, r_{t+1})$, to learn and improve its policy.

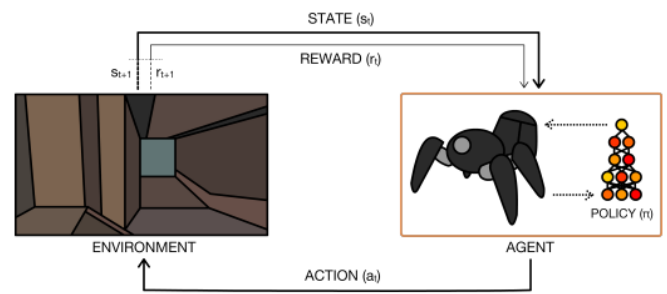

*Figure 2 Reinforcement Learning flow diagram example*

The dynamics of trial-and-error-learning has its roots in behaviourist psychology, being one of the main foundations of RL [31]. The best sequence of actions is determined by the rewards provided by the environment. Every time the environment transitions to a new state, it also provides a scalar reward $r_{t+1}$ to the agent as feedback. The goal of the agent is to learn policy π, which maximizes the expected return as a reward. For any given state, a policy returns an action to perform an optimal policy (any policy), that maximizes the expected return in the environment, RL aims to solve the problem of optimal control, where the agent needs to learn about the consequences of actions by trial-and-error [30]. To generate the rewards, the function uses the Bellman equation and takes two inputs, state and action.

$$R = \sum_{t=0}^{T-1} \gamma^t r_t(b_t, a_t)$$

*Equation 9 Deep Reinforcement Learning*

where **R** is the total return, **t** is the turn, **γ** is the discount factor, **r** is the reward, **b** stands for the dialogue state, and **a** stand for the action.

### 4.2.1 Q-Learning

Q-learning is an RL technique used in machine learning. "Q" names the function that returns the reward, used to provide the reinforcement and can be said to stand for the "quality" of an action taken in a given state [33]. Q-Learning can identify an optimal action-selection policy for given infinite exploration time and a partly-random policy. Q-Learning finds a policy that is optimal in the sense that it maximizes the expected value of the total reward over all successive steps, starting from the current state [34]. It is proven that, when the model is trained sufficiently under any policy, the algorithm converges with probability 1 to a close approximation of the action-value function for any target policy. Q-learning learns the optimal policy even when actions are selected according to a random policy [35]. These modified Q-values can be learned by a neural network. This is an iterative process of updating the values where the Q-function gives the better and better approximations by continuously updating the Q-values in the table. Q-Learning is a value-based reinforcement learning algorithm to find the optimal action-selection policy to maximize the value function Q. It helps maximize the expected reward by selecting the best of all possible actions. *Q(s, a)* returns the expected future reward of that action at that particular state, using epsilon greedy strategy.

$$Q(s, a) = Q(s, a) + \alpha (r + \gamma \max_{a'} Q(s', a') - Q(s, a))$$

*Equation 10 Deep Q-Learning (Deep Reinforcement Learning)*

where the first *Q(s, a)* stands for the new *Q* value on the left-hand side of the equation, which we need to calculate, the second *Q(s, a)* stands for the old *Q* value on the right-hand side of the equation, α stands for the learning rate (hyperparameter, initialized with the value 0.001; 0< α <=1; the value is decided based on the extent that how likely the newly acquired information overrides the old information), *r* is the reward, *γ* is the discount factor (to maximize the future sum of rewards and value initialized at 1), the $\max_{a'} Q(s', a') - Q(s, a)$ suggest the estimation of optimal future value, and $(r + \gamma \max_{a'} Q(s', a') - Q(s, a))$ represents the learned value.

### 4.2.2 Q-Learning Algorithm

| 1 | Initialize *Q (s, a)* |
|---|---|
| 2 | Repeat (for each interaction): |
| 3 |   Initialize *s* |
| 4 |   Repeat (for each step of interaction): |
| 5 |     Choose *a* from *s* using policy derived from *Q* |
| 6 |     Take action *a*, observe *r, s'* |
| 7 |     $Q(s, a) = Q(s, a) + \alpha (r + \gamma \max_{a'} Q(s', a') - Q(s, a))$ |
| 8 |     $S \leftarrow s'$ |
| 9 | Until *s* is terminal |

*Table 1 Algorithm for Q-Learning*

where [35],

**α** : the learning rate set between 0 and 1. Setting it to 0 means that Q-values are never updated hence nothing is learned. Setting a high value such as 0.9 means that learning can occur quickly.

**γ** : the discount factor also set between 0 and 1. This represents that, the future rewards are worth less than immediate rewards. The discount factor needs to be less than 0 for the algorithm to converge.

$\max_{a'}$ : the maximum reward that is, attainable in the state following the current one.

### 4.3 POMDP with updated BS history to improve POLICIES

Considering what is stated by the thesis work [21], 7 tuples are used such as, states, actions, observations, transition, observation probability, reward functions, discount factor, etc. this approach has stated the policy selection based on knowledge level selection among different modes, Expert; Professional; Amateur; Novice, the policy selector module get the mode as input and return respective policy as output. Policies are mapped from belief-state values to action. If we receive the minimum sharp points and the maximum sharp point as high as possible for the EXPERT level that will be counted as the optimized length of dialogue and higher accuracy of policy. But the work shows that over a few datasets, the results are not so good as NOVICE level has a higher minimum and maximum sharp points, so that can be worked out. The following architecture in **Error! Reference source not found.** shows how knowledge level threshold are selected using BSH.

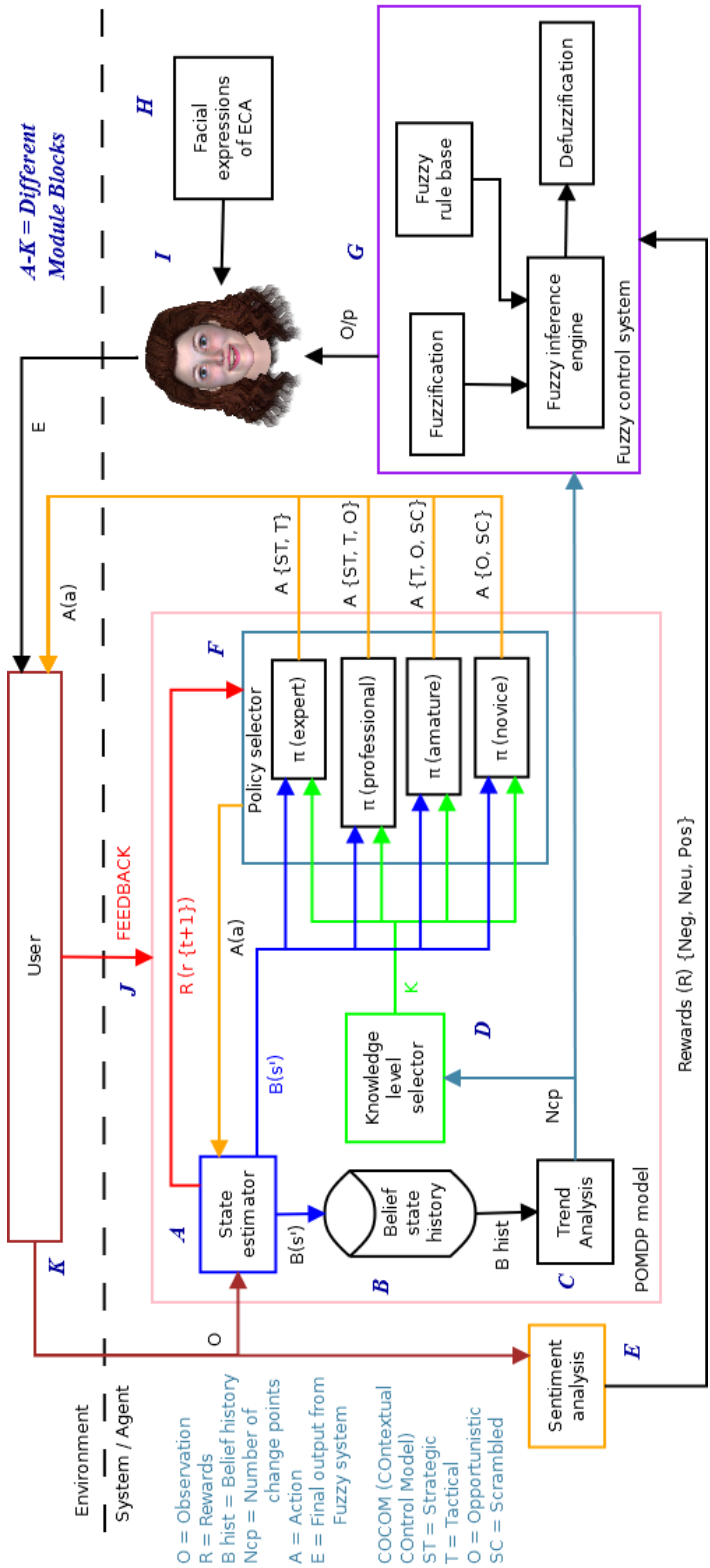

*Figure 3 Proposed POMDP-based Dialogue Management incorporating sentiment analysis and Reinforcement Learning architecture*

### 4.4 Algorithm

| 1 | isGoalState ← false |
|---|---|
| 2 | belief ← 1 |
| 3 | CREATE empty LIST bhist |
| 4 | ADD belief to LIST bhist |
| 5 | WHILE isGoalState NOT EQUAL true |
| 6 |   input ← READ(observation) |
| 7 |   IF input EQUAL 'exit' THEN |
| 8 |     isGoalState ← true |
| 9 |   ELSE |
| 10 |     b(s') ← StateEstimator(input, belief) |
| 11 |     Rewards (R(r{t+1})) ← StateEstimator(input, belief) |
| 12 |     π{m} ← Rewards (R(r{t+1})) |
| 13 |     ADD b(s') to LIST bhist |
| 14 |     Ncp ← TrendAnalysis(bhist) |
| 15 |     k ← KnowledgeLevelSelector(Ncp) |
| 16 |     R ← SentimentAnalysis{Neg, Neu, Pos} |
| 17 |     Ncp ← {ST, T, O, SC} |
| 18 |     π{m} ← PolicySelector(k) |
| 19 |     action ← MakeAction(π{m}, b(s')) |
| 20 |     belief ← b(s') |
| 21 |     PRINT action |
| 22 |   ENDIF |
| 23 | END WHILE |

*Table 2 System Algorithm*

## 5. Implementation and Experiment setup

### 5.1 User Interface (UI)

Different modules in the experimental setup application,

- First-half - 3D model, to discover the intention; to detect the emotion, and to display the emotion over the 3D model.
- Second half - Dialogue Management (DM) component where interaction between the user and an agent happens
- The edit text and three buttons - an Edit Text to write textual messages, first button - send button for textual-message, second button - record button for live detection emotion from the speech (voice), third button - record button to record speech to convert it to textual based messages on DM component
- Second half during the live recording session - It represents the probability of emotion which is detected from the voice at the user's end

The working prototype (proof of concept) can be found as a video here[1] and another example with different thought is found as a video is here[2]. A tiny video which shows the lip syncing with the speech on ECA model can be found here[3]. A speech recording, sentiment discovering demo can be found here[4].

### 5.2 Datasets

For the basic setup, the service data are used which was given by Mulpuri [21] and is available at the link[5]. To improve intentions and to improve the emotion detection accuracy, I have incorporated different datasets from different sources. The full sentence-based emotion detection can be found at the link[6], which has 7000+ *lines* of conversation to train the model and to track the exact emotion, 1500+ emotion-related *words* were gathered which can be found at the link[7].

### 5.3 User-Agent conversational chat setup

The conversation setup between the user and the agent was developed on python platform. The obtained results as in Appendix A: User-Agent chat conversation Ontology-based customization: Online Book Search portal are represented in the sentiments analysis rewards as in Appendix B: User-Agent chat conversation: Sentiment Analysis rewards for *Online book search portal customization* example. The Python, Android codes are available here[8], but a person seeking a grant need to make a request when redirected to the Google drive page, then after your request is granted, the codes will be accessible.

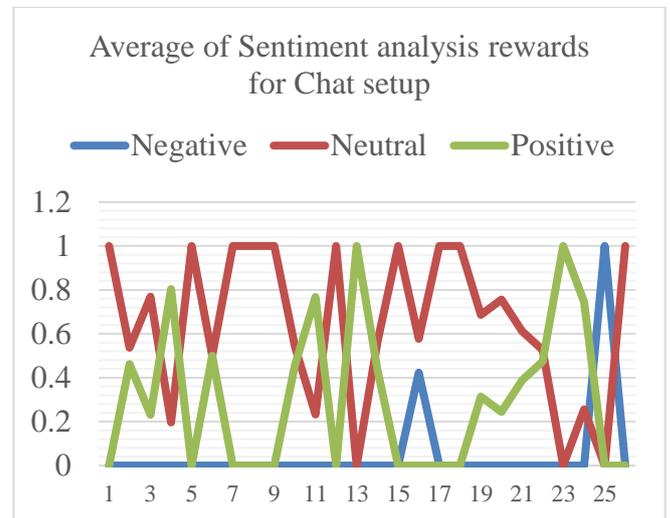

*Figure 4 Rewards chart: Chat conversation setup*

## 6. Simulation and Results

From the basic setup, 10 tests have been executed but the result of only 1 test is shown as in the link[9]. These outputs are gathered from the Log data of the Android Studio IDE. As you can see, the knowledge level is Amateur. By knowing the knowledge level, we can get the belief state history and the policy. These results are obtained on 3 different machines using the Android Studio IDE. The 3 devices which are tested are listed below.

- Real Android Samsung A7 device
- Genymotion simulator
- Android Studio ADB Emulator, etc.

### 6.1 Simulation design and process

- Design- Trained the knowledge level selectors with ontology and hand-crafted policies; Performing trend analysis over BSH to find Ncp; Pretended to be the different user each time while maintaining the fluctuations to categorize four different types of users using Ncp; Ncp is used to set the ground truth for the policy generation
- Process- For the training part in the knowledge level selectors; the technological and knowledge level of the user should be used as one of the aspects when analysing the policy (from business

---

[1] https://youtu.be/UbjucsBzTgU

[2] https://youtu.be/z6F0yO_4UM8

[3] https://youtu.be/lHHRbfcAOAo

[4] https://youtu.be/CND_J32A3MM

[5] https://drive.google.com/open?id=1S0rjk-nW_yanIxQ6S_obF9LVKXE4waVq

[6] https://drive.google.com/open?id=1XQ3dfXrHPpSGRUM0uW83M6z0b5olIW27

[7] https://drive.google.com/open?id=16ucMMiLbucFgn0EDM8ZtdyA1awqsRNyl

[8] https://drive.google.com/open?id=19DU290wo3i-x5QsMEMpE3kXjthf1ZnMj

[9] https://drive.google.com/open?id=1r9c1Sj8rUiVZ3vYLXaCt5q36YaBkX8fm

- management case study[10]); Ncp is used to simulate the chat conversation
- Ground truth- while training, the threshold is set for different users as *n% (n= 0 to 100)* of times Ncp found. 0-25%- Expert | 25-50%- Professional | 50-75%- Amateur | 75-100%- Novice. If Ncp falls into any such category of threshold, then that becomes the policy for the user, for that loop of interaction. Then, the simulated results are compared with the hand-crafted policy and set threshold to match whether policy is improved or not!

Based on the interaction and different iterations of the user-agent conversation in Appendix A: User-Agent chat conversation Ontology-based customization: Online Book Search portal, the belief state history database is recorded and then the trend analysis is performed over it, so that sharp points (minimum and maximum) can be found.

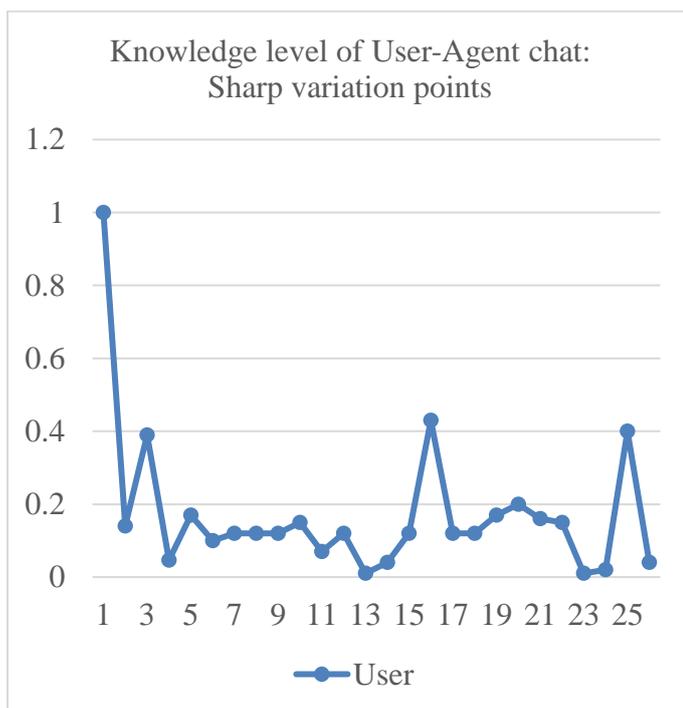

*Figure 5 Knowledge users: Sharp variation Points*

The wavelet analysis transforms the signal and represents the signal using positive or negative values. A zero-crossing point between the values of detail coefficients represents the sharp variation points.

---

[10] https://drive.google.com/open?id=1240QQp-ERDJcApVlWJJocfl2zFS1xhfK

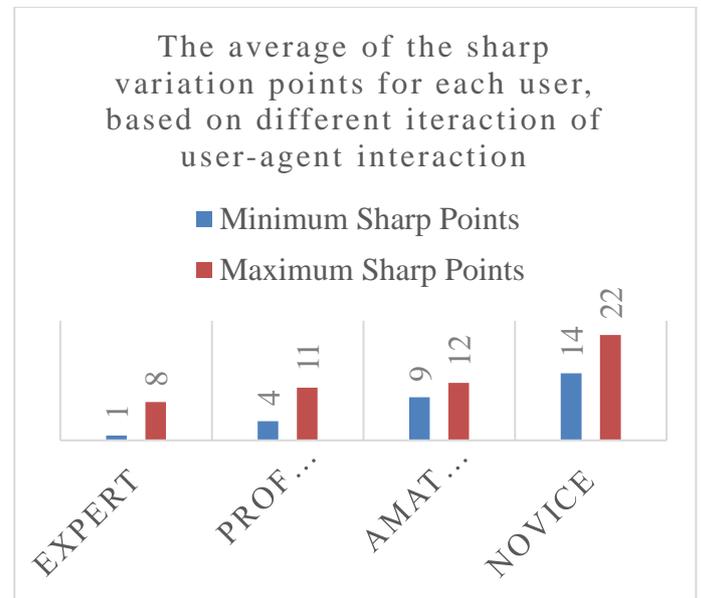

*Figure 6 Trend Analysis- Sharp Variation Points chart*

The matrix values can be calculated from the defuzzification step of the Fuzzy Logic System. The work of Fuzzy logic system was brought by the thesis by Kaur [38] and I am using those formulas to calculate the **Wn** value, where **n** varies from 1 to 12, for FAMM table. The equation for the same is given below.

$$X = \frac{\sum W_n * FAMM_n}{\sum W_n}$$

*Equation 11 Defuzzification to generate crisp output*

| Fear | Sad |
|---|---|
| 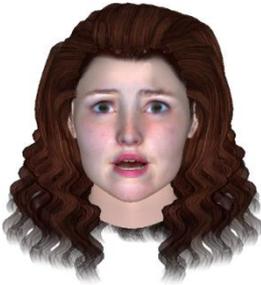 | 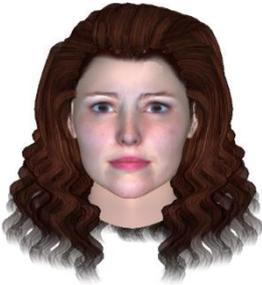 |

*Table 3 Final ECA emotion after crisp Fuzzy Logic System output*

The Q-Learning rewards are as given in the figure below which affects the decision-making process.

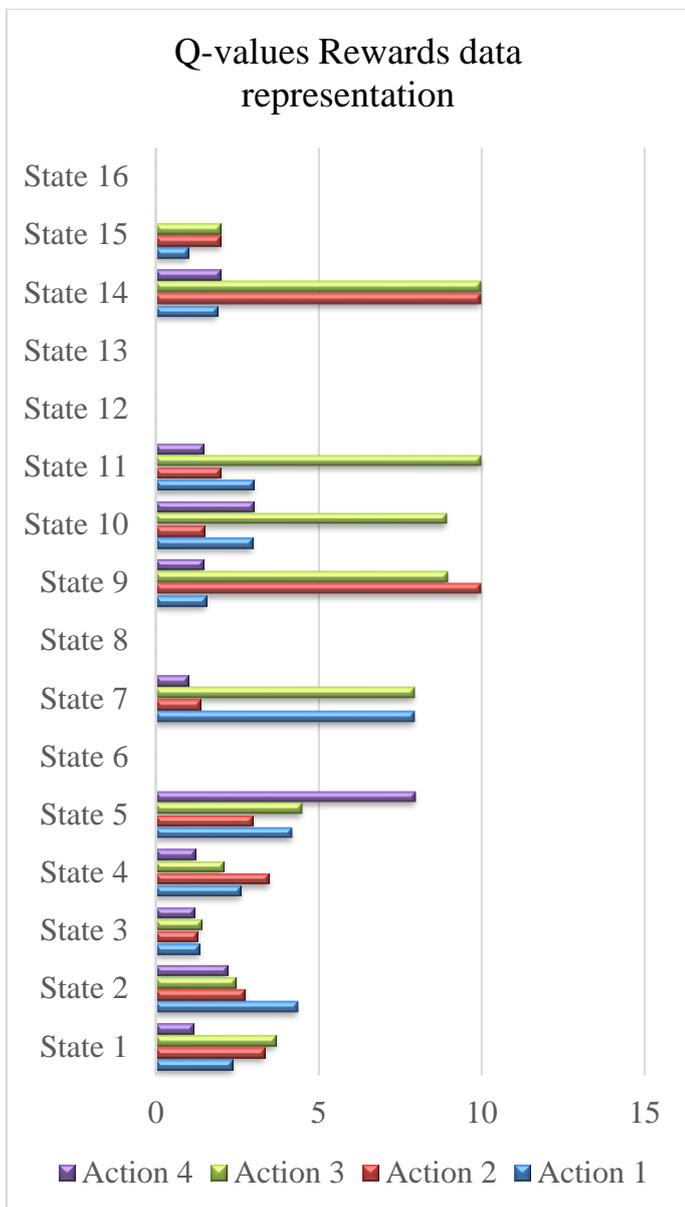

*Figure 7 Q-Values chart for all users*

We choose the action a and state s based on the Q-table we get. Initially every Q-value starts at 0. In the beginning as explained the epsilon rates will be higher until the exploration happens and suitable but random action is chosen, the logic is the agent doesn't know anything as such in terms of handling the uncertainty. As agent explores more, the rate decreases, which offers the 0 value until the last interaction with a few fluctuations in between as well, if the confidence measure fluctuates. The perceptron-action-learning loop for each interaction, provides the Q-values to get the new state s' based on the action a made using the reward r.

### 6.2 Policy improvement discussion

The policies were hand-crafted during the knowledge level training for the simulation purpose. Then the goal is to improve the policy using the self-optimization while the conversation last.

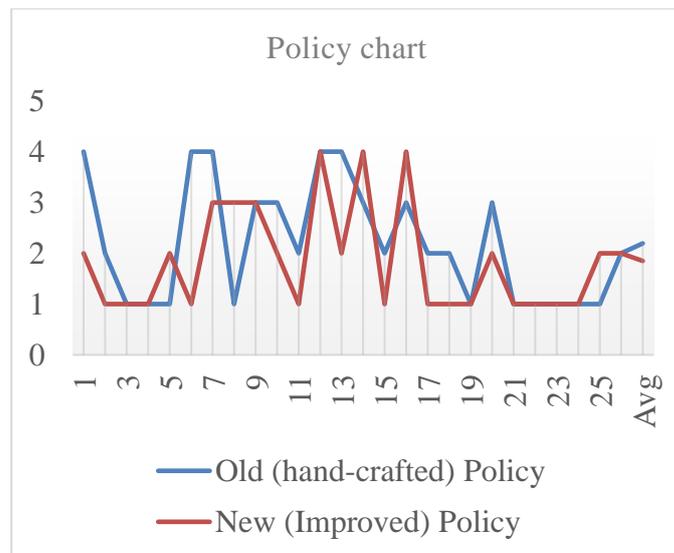

*Figure 8 Policy chart*

Henceforth, having the intention discovery for each interaction iteration, action is made by the agent in terms of policy, authenticated by the user. The policy chart represents the difference between the old and the new improved policy based on the data given in **Error! Reference source not found.**. Where, 1=Expert; 2=Professional; 3=Amateur; 4=Novice. As shown, the last point represents the average of all the policies and it is proven that, Policy has been improved.

### 6.3 Results and Accuracy discussion

We can analyse the results to see if the goal of the paper was achieved, in terms of accuracy; dialogue length, improvement in the ECA emotion. The accuracy is calculated on average accuracy of 4 different kinds of knowledge levels which are, Expert, Professional, Amateur, and Novice; where the goal was to reach certain decision to resolve user's goal in the decision-making process. Average dialogue length is measured using the different iterations over different 4 knowledge levels of the users. The improvement in the results of an ECA emotion can be concluded based on the user expressing fewer negative emotions and more neutral and positive emotions generation, as users will not accommodate the frequent anger; sad; fear; etc. emotions, therefore sentiment analysis helps in determining the more neutral when not the positive to leave enhancing impact on the user.

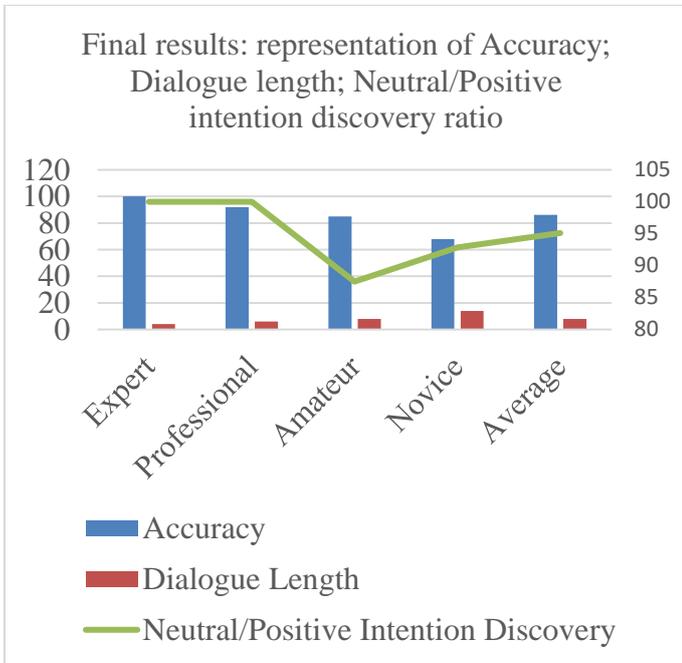

*Figure 9 Accuracy calculation*

The accuracy is calculated based on the interaction iteration that how many times user was able to reach the goal and how long it took (dialogue length).

| Users | Accuracy (in %) | Dialogue Length (in units) | Neutral/Positive Intention Discovery (in %) |
|---|---|---|---|
| Expert | 100 | 4 | 100 |
| Professional | 92 | 6 | 100 |
| Amateur | 85 | 8 | 87.5 |
| Novice | 68 | 14 | 92.86 |
| *Average* | **86.25** | **8** | **95.09** |

*Table 4 Results: Accuracy; Dialogue length; Neutral/Positive intention discovery ratio*

| Case | Bui's results | Mulpuri's results | Kaur's results | This paper's results |
|---|---|---|---|---|
| *Accuracy* | 67.9% | 79.3% | N/A | 86.25% (based on the average of all 4 types of knowledge level users) |
| *Average Dialogue length* | 34 | 21 | N/A | 8 (worst case: 14) (average of different interaction iterations of all 4 types of users) |
| *ECA emotion improvement* | N/A | N/A | Conducted usability study to test enhanced user's experience | Used sentiment analysis and Fuzzy crisp output to generate emotions based on the rewards and results show that **95%** of the time neutral or positive emotions are generated |

*Table 5 Final results discussion*

## 6.4 Discussion

When we traverse through different knowledge levels of policy selection, there is a fall in the accuracy measurements starting from expert to professional to amateur to novice. But, to note in the case of dialogue length it is complete opposite as numbers of dialogue increases for the same traversing path. That way conclusion can be made that, accuracy and the dialogue length are proportional to each other. The intention discovery improvement ratio is calculated based on number of times positive or neutral emotions were generated.

## 6.5 Limitations

There are still challenges, user input in slang (lingo) will not offer good accuracy to reach optimal policy, as sentiment cannot be detected and misunderstanding might occur (to overcome this limitation, training can be done such a way that regional-slangs are understood); RL operates on trial-and-error while solving the optimal control aim, which generates uncertain consequences as actions in the environment are still under training; the optimal policy must be inferred by trial-and-error interaction with the environment, the only learning signal the agent receives is the reward; the observations of the agent depend on its actions and can contain strong temporal correlations; agents must deal with long-range time dependencies to handle the consequences of action which only materialize after many iterations of the environment which is known as a credit assignment problem.

## 7. Conclusion and Future work

### 7.1 Conclusion

Based upon the research, ECA is currently believed to be the best efficient medium for HCI. POMDP outperforms all the DM based domain's approaches under the uncertainty and under the stochastic environment, using the knowledge level selectors which enhances the higher threshold out of training to get the optimal policy and its following action. To optimize the dialogue length, traditional approaches fail miserably as they are not intelligent enough to compete with the uncertainty present in human conversation. POMDP-based DM helps in improving intention discovery for ECA and eventually helps in improving policy using the Q-Learning rewards and sentiment analysis helps in understanding emotion detection from user input to decide goal-driven aim achievement with the help of human-like emotions on the 3D model face; where the rewards are considered and converted to the emotions using the fuzzy logic system which produces the crisp emotion using COCOM. Reinforcement Learning helps in learning optimal policy which helps in reducing dialogue length, making dialogue conversation smooth than the former.

### 7.2 Future work

- SDS can be incorporated on the future POMDP models.
- The best possible training datasets with over one million dialogues along with hand-crafted policy which can train supervised knowledge level selection for each single dialogue to solve credit assignment problem: requires man-power and an investment, can be referred to train the model to improve of the policy in the Reinforcement Learning context. That helps in emotion detection in improving on a large scale to handle the unexpected uncertainty.
- Discovery of user intention will be the essential part of the work to enhance the policy selection over knowledge threshold which eventually satisfies the user, but the knowledge level training can be made such a way that, the domain-specific customization doesn't affect the emotion detection from user and representing at agent's end on 3D model.
- If possible, the full-fledged working commercial application can be developed from the proof of concept developed as part of this thesis. The policy should be trained using huge datasets to improve the scalability.
- Some limitations such as, sarcasm detection, slangs, can be trained to be handled with plethora of word embeddings and the relevance.
- Word embeddings can be visualized in a stochastic environment to plot the similar wordings in a corpus to cluster them, which can enhance the limitation of language-accent-barrier; locale-regional barrier.
- The audio can be improved as there is still a scope to make it sound like more humanish than the current robotic sound. Apart from that, verbal communication can be improved to make ECA lip sync the conversation with the printed text.

# Appendices

Appendix A: User-Agent chat conversation Ontology-based customization: Online Book Search portal

Where, U: User; A: Agent; n: 1 to 26 (dialogue iterations)

| n | | |
|---|---|---|
| | A: | Hi, I am AVATAR. Welcome to use the interactive software customization tool. The functionality of Search a book is essential and it has been added as a required feature. As for the way of searching a book, how would you like your system to provide details? |
| 1 | U: | May I know what do you mean by providing details? |
| | A: | Functionality get detailed info of a book is a sub-requirement of searching a book. Do you need it? |
| 2 | U: | Yes i would love to know more about searching a book functionality. |
| | A: | When you input the id of a book, get detailed info of a book, returns detailed information about the book, like publication info and contents. It has sub-requirements. |
| 3 | U: | Yes, i prefer my system to provide detailed information of book. |
| | A: | As this optional feature requires the functionality of get detailed info of a book, they have also been added to your system. |
| 4 | U: | That's great. |
| | A: | How would you like your system to allow user to search for a book? |
| 5 | U: | It should be simply a keyword-based searching. |
| | A: | To display the searching result, do you |

| | | | | | |
|---|---|---|---|---|---|
| | | want the list of books being sorted? | 17 | U: | I guess, i will opt in for exact match |
| 6 | U: | Yes please add sort the books functionality. | | A: | Do you want to add, manage shopping cart functionality? |
| | A: | Functionality pick a book from a list has already been selected due to the decision on search a book. | 18 | U: | What else do i get in cart functionality? |
| | | | | A: | You get to add book to a cart, remove from a cart and list all the items in a cart. |
| 7 | U: | What else is included apart from pick a book under search a book functionality? | 19 | U: | Okay i want to have the functionality of managing cart. |
| | A: | We offer quick view; detailed view; and sorting by title or latest or popular. | | A: | Okay sure. Do you want to add the payment gateway? |
| 8 | U: | Which other searching functionality do you offer? | 20 | U: | What is payment gateway and how can it be useful? |
| | A: | We have basic search and advanced search and search by author or title or publication functionality. | | A: | Functionality payment gateway comes with high and low security to receive online payment for the book ordered online. |
| 9 | U: | And what do you offer in keyword matching module? | 21 | U: | Okay add payment gateway. |
| | A: | We have broad match and exact match. That will be added by default as you have added for keyword-based searching functionality. | | A: | Functionality get summary is part of place an order functionality, would you like to add? |
| 10 | U: | Ok i understand. How can i add advanced search? | 22 | U: | Yes add summary functionality |
| | | | | A: | Do you want to add set delivery address functionality? |
| | A: | Functionality advanced search is a sub-requirement of searching a book. Do you want it? | 23 | U: | Yes. |
| | | | | A: | Okay. Set payment information functionality has been added under payment gateway functionality. |
| 11 | U: | Sure. Please add advanced search. | | | |
| | A: | The quality constraint broad match is related to search in book keywords. Do you need it? | 24 | U: | That's good. |
| | | | | A: | All the requirements have been evaluated. Would you like to make any changes? |
| 12 | U: | What? | | | |
| | A: | Your response cannot be recognized. Please answer with the suggested response. | 25 | U: | No. |
| | | | | A: | The customization process is complete. Thank you for your cooperation. |
| 13 | U: | Okay. | | | |
| | A: | The quality constraint broad match is related to search in book keywords. Do you need it? | 26 | U: | Quit |
| | | | | A: | Thank you and see you soon. |
| 14 | U: | Can you please explain? | | | |
| | A: | When you search a book, not only books with keywords exactly like your inputs will be returned, but books with keywords like your inputs will also be returned. If you choose constraint broad match, quality constraint exact match cannot be selected. | | | |
| 15 | U: | Now it makes sense. | | | |
| | A: | Do you want broad match? | | | |
| 16 | U: | No, i don't think so. | | | |
| | A: | The quality constraint exact match is related to search in book keywords. Do you need it? | | | |

*Table 6 Chat conversation setup User-Agent*

Appendix B: User-Agent chat conversation: Sentiment Analysis rewards

| n | Sentiment rewards |
|---|---|
| 1 | May i know what do you mean by providing details? *Compound: 0.0, neg: 0.0, neu: 1.0, pos: 0.0,* |
| 2 | Yes i would love to know more about searching a book functionality. *Compound: 0.7845, neg: 0.0, neu: 0.537, pos: 0.463,* |
| 3 | Yes, i prefer my system to provide detailed information of book. |
| 4 | *Compound: 0.4019, neg: 0.0, neu: 0.769, pos: 0.231,*<br>That's great.<br>*Compound: 0.6249, neg: 0.0, neu: 0.196, pos: 0.804,* |
| 5 | It should be simply a keyword-based searching.<br>*Compound: 0.0, neg: 0.0, neu: 1.0, pos: 0.0* |
| 6 | Yes please add sort the books functionality.<br>*Compound: 0.6124, neg: 0.0, neu: 0.5, pos: 0.5,* |
| 7 | What else is included apart from pick a book under search a book functionality?<br>*Compound: 0.0, neg: 0.0, neu: 1.0, pos: 0.0,* |
| 8 | Which other searching functionality do you offer?<br>*Compound: 0.0, neg: 0.0, neu: 1.0, pos: 0.0,* |
| 9 | And what do you offer in keyword matching module?<br>*Compound: 0.0, neg: 0.0, neu: 1.0, pos: 0.0,* |
| 10 | Ok i understand. How can i add advanced search?<br>*Compound: 0.4939, neg: 0.0, neu: 0.543, pos: 0.457,* |
| 11 | Sure. Please add advanced search.<br>*Compound: 0.6808, neg: 0.0, neu: 0.233, pos: 0.767,* |
| 12 | What?<br>*Compound: 0.0, neg: 0.0, neu: 1.0, pos: 0.0,* |
| 13 | Okay.<br>*Compound: 0.2263, neg: 0.0, neu: 0.0, pos: 1.0,* |
| 14 | Can you please explain?<br>*Compound: 0.3182, neg: 0.0, neu: 0.566, pos: 0.434,* |
| 15 | Now it makes sense.<br>*Compound: 0.0, neg: 0.0, neu: 1.0, pos: 0.0,* |
| 16 | No, i don't think so.<br>*Compound: -0.296, neg: 0.423, neu: 0.577, pos: 0.0,* |
| 17 | I guess, i will opt in for exact match<br>*Compound: 0.0, neg: 0.0, neu: 1.0, pos: 0.0,* |
| 18 | What else do i get in cart functionality?<br>*Compound: 0.0, neg: 0.0, neu: 1.0, pos: 0.0,* |
| 19 | Okay i want to have the functionality of managing cart.<br>*Compound: 0.296, neg: 0.0, neu: 0.686, pos: 0.314,* |
| 20 | What is payment gateway and how can it be useful?<br>*Compound: 0.4404, neg: 0.0, neu: 0.756, pos: 0.244,* |
| 21 | Okay add payment gateway.<br>*Compound: 0.2263, neg: 0.0, neu: 0.612, pos: 0.388,* |
| 22 | Yes add summary functionality<br>*Compound: 0.4019, neg: 0.0, neu: 0.526, pos: 0.474,* |
| 23 | Yes.<br>*Compound: 0.4019, neg: 0.0, neu: 0.0, pos: 1.0,* |
| 24 | That's good.<br>*Compound: 0.4404, neg: 0.0, neu: 0.256, pos: 0.744,* |
| 25 | No.<br>*Compound: -0.296, neg: 1.0, neu: 0.0, pos: 0.0,* |
| 26 | Quit<br>*Compound: 0.0, neg: 0.0, neu: 1.0, pos: 0.0,* |

*Table 7 User-Agent interaction: Sentiment Analysis Rewards*